\documentclass[10pt,twocolumn,letterpaper]{article}

\usepackage[pagenumbers]{cvpr}

\usepackage{graphicx}
\usepackage{amsmath}
\usepackage{amssymb}
\usepackage{booktabs}

\usepackage{algorithm}
\usepackage{algorithmic}
\usepackage{multirow}
\usepackage{bbding}
\usepackage{bbm}
\usepackage{enumitem}
\usepackage{xcolor}
\usepackage{colortbl} % using color table
\definecolor{citecolor}{HTML}{0071bc}
\usepackage[pagebackref,breaklinks,colorlinks, anchorcolor=blue, citecolor=citecolor]{hyperref}

\usepackage{multirow}

\usepackage[capitalize]{cleveref}
\crefname{section}{Sec.}{Secs.}
\Crefname{section}{Section}{Sections}
\Crefname{table}{Table}{Tables}
\crefname{table}{Tab.}{Tabs.}

\def\eg{\emph{e.g}\onedot}

\newcommand*\samethanks[1][\value{footnote}]{\footnotemark[#1]}

\begin{document}


\title{Instance-specific and Model-adaptive Supervision for \\ Semi-supervised Semantic Segmentation}

\author{Zhen Zhao\textsuperscript{\rm 1,2}\thanks{Equal contribution.}~~~~Sifan Long\textsuperscript{\rm 3,1}\samethanks~~~~Jimin Pi\textsuperscript{\rm 2}~~~~Jingdong Wang\textsuperscript{\rm 2}\thanks{Corresponding authors.}~~~~Luping Zhou\textsuperscript{\rm 1}\samethanks \\
\textsuperscript{\rm 1}University of Sydney\hspace{16mm}
\textsuperscript{\rm 2}Baidu VIS\hspace{16mm}
\textsuperscript{\rm 3}Jilin University\\
% {\tt\small \{zhen.zhao, luping.zhou\}@sydney.edu.au}\\
% {\tt\small summitlsf@outlook.com~~\{pijimin01, wangjingdong\}@baidu.com}
}
\maketitle

%%%%%%%%% ABSTRACT
\begin{abstract}
Recently, semi-supervised semantic segmentation has achieved promising performance with a small fraction of labeled data. However, most existing studies treat all unlabeled data equally and barely consider the differences and training difficulties among unlabeled instances. Differentiating unlabeled instances can promote instance-specific supervision to adapt to the model's evolution dynamically.
In this paper, we emphasize the cruciality of instance differences and propose an instance-specific and model-adaptive supervision for semi-supervised semantic segmentation, named \textbf{iMAS}.
Relying on the model's performance, iMAS employs a class-weighted symmetric intersection-over-union to evaluate quantitative hardness of each unlabeled instance and supervises the training on unlabeled data in a model-adaptive manner. Specifically, iMAS learns from unlabeled instances progressively by weighing their corresponding consistency losses based on the evaluated hardness. Besides, iMAS dynamically adjusts the augmentation for each instance such that the distortion degree of augmented instances is adapted to the model's generalization capability across the training course.
Not integrating additional losses and training procedures, iMAS can obtain remarkable performance gains against current state-of-the-art approaches on segmentation benchmarks under different semi-supervised partition protocols\footnote{Code and logs: \url{https://github.com/zhenzhao/iMAS}.}.
\end{abstract}

%%%%%%%%%%%%%%%%%%%%%%%%%%%%%%%%%%%%
%%  1. introduction
%%%%%%%%%%%%%%%%%%%%%%%%%%%%%%%%%%%%
\section{Introduction}
\label{sec:intro}

Though semantic segmentation studies~\cite{seg15fcn,seg18deeplabv3plus} have achieved significant progress, their enormous success relies on large datasets with high-quality pixel-level annotations. 
Semi-supervised semantic segmentation (SSS)~\cite{sss18advseg,sss19s3gan} has been proposed as a powerful solution to mitigate the requirement for labeled data.
Recent research on SSS has two main branches, including the self-training (ST)~\cite{ssl13pseudo} and consistency regularization (CR)~\cite{ssl17mt} based approaches.
\cite{sss22st++} follows a self-training paradigm and performs a selective re-training scheme to train on labeled and unlabeled data alternatively.
Differently, CR-based works \cite{sss20cct,sss22PSMT} tend to apply data or model perturbations and enforce the prediction consistency between two differently-perturbed views for unlabeled data. 
In both branches, recent research  \cite{sss20cutmixseg,sss21simple,sss21ael}  demonstrates that strong data perturbations like CutMix can significantly benefit the SSS training. To further improve the SSS performance, current state-of-the-art approaches~\cite{sss21semisegcontrast,sss22u2pl} integrate the advanced contrastive learning techniques into the CR-based approaches to exploit the unlabeled data more efficiently. Works in \cite{sss20scn,sss22ELN} also aim to rectify the pseudo-labels through training an additional correcting network.

Despite their promising performance, SSS studies along this line come at the \textbf{cost} of introducing extra network components or additional training procedures. In addition, majorities of them treat unlabeled data equally and completely ignore the differences and learning difficulties among unlabeled samples. 
For instance, \textbf{randomly and indiscriminately} perturbing unlabeled data can inevitably over-perturb some difficult-to-train instances. Such over-perturbations
exceed the generalization capability of the model and  hinder effective learning from unlabeled data.
As discussed in~\cite{sss21simple}, it may also hurt the data distribution.
Moreover, in most SSS studies, final consistency losses on different unlabeled instances are minimized in an \textbf{average}  manner. However, blindly averaging can implicitly emphasize some difficult-to-train instances and result in model overfitting to noisy supervision.

In this paper, we emphasize the cruciality of instance differences and aim to provide instance-specific supervision on unlabeled data in a model-adaptive way. There naturally exists two main questions. First, how can we differentiate unlabeled samples? We design an instantaneous instance ``hardness," to estimate 1) the current generalization ability of the model and 2) the current training difficulties of distinct unlabeled samples. Its evaluation is closely related to the training status of the model, \textit{e.g.,} a difficult-to-train sample can become easier with the evolution of the model. Second, how can we inject such discriminative information into the SSS procedure? Since the hardness is assessed based on the model's performance, we can leverage such information to adjust the two critical operations in SSS, \textit{i.e.,} data perturbations and unsupervised loss evaluations, to adapt to the training state of the model dynamically.

\begin{figure*}[t]
\centering
\vspace{-2em}
\includegraphics[width=0.86\linewidth]{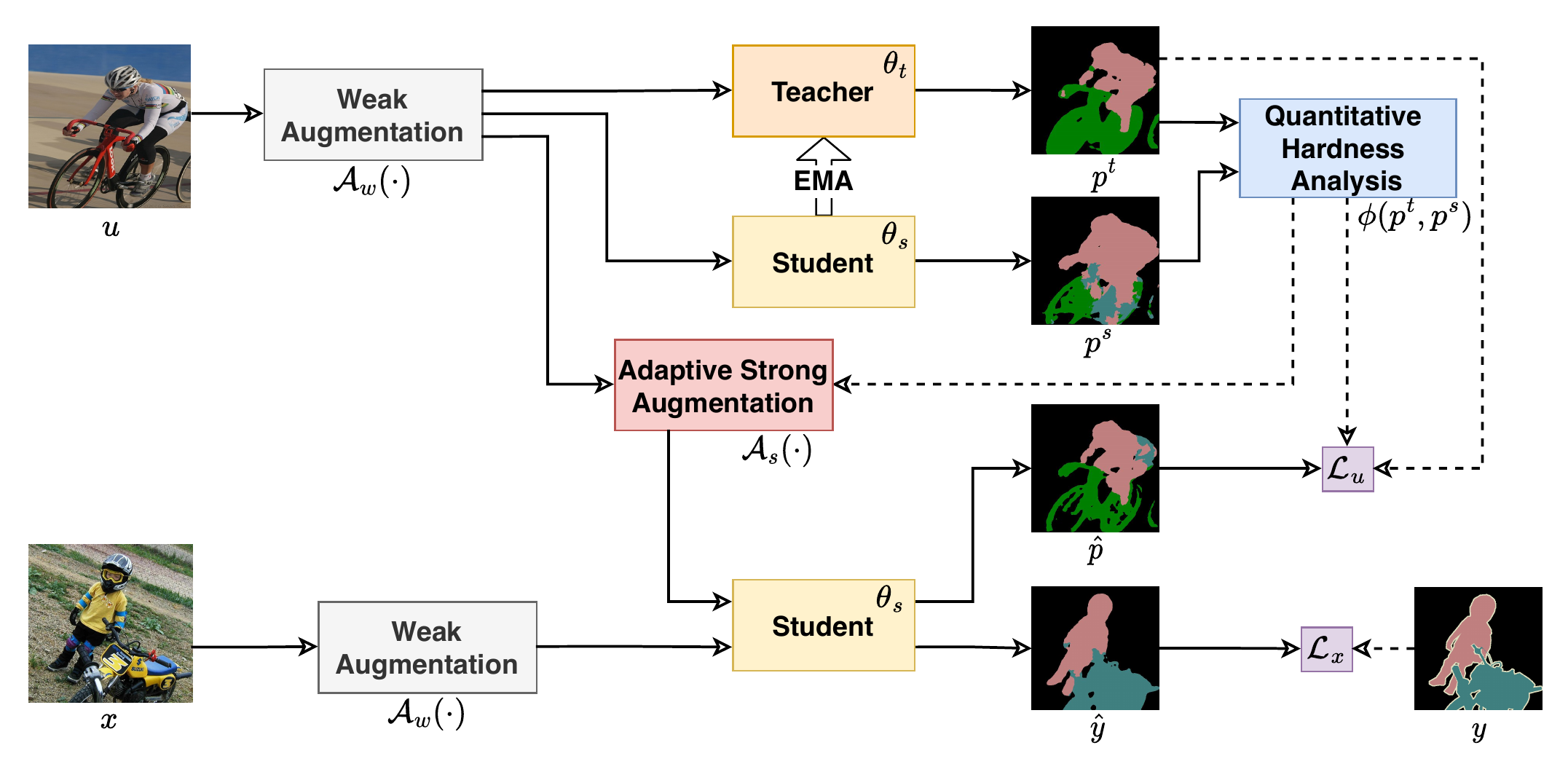}
\caption{Diagram of our proposed iMAS. 
In a teacher-student framework, labeled data $(x,y)$ is used to train the student model, parameterized by $\theta_s$, by minimizing the supervised loss $\mathcal{L}_x$. Unlabeled data $u$, weakly augmented by $\mathcal{A}_w(\cdot)$, is first fed into both the student and teacher models to obtain predictions $p^s$ and $p^t$, respectively. Then we perform quantitative hardness evaluation on each unlabeled instance by strategy $\phi(p^t, p^s)$. Such hardness information can be subsequently utilized: 1) to apply an adaptive augmentation, denoted by $\mathcal{A}_s(\cdot)$, on unlabeled data to obtain the student model's prediction $\hat{p}$; 2) to weigh the unsupervised loss $\mathcal{L}_u$  in a instance-specific manner. The teacher model's weight, $\theta_t$, is updated by the exponential moving average (EMA) of $\theta_s$ across the training course. }
\label{fig:diagram}
\vspace{-0.4em}
\end{figure*}

Motivated by all these observations, we propose an instance-specific and model-adaptive supervision, named \textbf{iMAS}, for semi-supervised semantic segmentation. As shown in \Cref{fig:diagram}, following a standard consistency regularization framework,  iMAS jointly trains the student and teacher models in a mutually-beneficial manner. The teacher model is an ensemble of historical student models and generates stable pseudo-labels for unlabeled data. 
Inspired by empirical and mathematical analysis in \cite{grill2020bootstrap,tian2021understanding}, difficult-to-train instances may undergo considerable disagreement between predictions of the EMA teacher and the current student. Thus in iMAS, we first evaluate the instance hardness of each unlabeled sample by calculating the class-weighted symmetric intersection-over-union (IoU) between the segmentation predictions of the teacher (the historical) and student (the most recent) models. Then based on the evaluation, we perform model-adaptive data perturbations on each unlabeled instance and minimize an instance-specific weighted consistency loss to train models in a curriculum-like manner. In this way, different unlabeled instances are perturbed and weighted in a dynamic fashion, which can better adapt to the model's generalization capability throughout the training processes. 

Benefiting from this instance-specific and model-adaptive design, iMAS obtains state-of-the-art (SOTA) performance on Pascal VOC 2012 and Cityscapes datasets under different partition protocols. For example, our method obtains a high mIOU of 75.3\% with only 183 labeled data on VOC 2012, which is 17.8\% higher than the supervised baseline and 4.3\% higher than the previous SOTA. Our main contributions are summarized as follows,
\begin{itemize} [leftmargin=*]
    \item iMAS can boost the SSS performance by highlighting the instance differences, without introducing extra network components or training losses.
    \item We perform a quantitative hardness-evaluating analysis for unlabeled instances in segmentation tasks, based on the class-weighted teacher-student symmetric IoU.
    \item We propose an instance-specific and model-adaptive SSS framework that injects instance hardness into loss evaluation and data perturbation to dynamically adapt to the model's evolution.
    % dynamically.
\end{itemize}

%%%%%%%%%%%%%%%%%%%%%%%%%%%%%%%%%%%%
%%  2. related works
%%%%%%%%%%%%%%%%%%%%%%%%%%%%%%%%%%%%
\section{Related work}
\label{sec:rwork}

Recent studies on CR-based semi-supervised learning have achieved impressive improvements in classification tasks~\cite{semi2}. Based on clustering assumptions, these methods enforce prediction consistency on the unlabeled sample with different perturbations. Early works like Mean-Teacher~\cite{ssl17mt} aimed to generate a more robust and accurate pseudo-label using ensemble techniques. VAT~\cite{ssl18vat}, UDA~\cite{ssl20uda}, and MixMatch~\cite{ssl19mixmatch} then improved the performance by using more advanced augmentations, like adversarial perturbations~\cite{goodfellow2014explaining}, randomAug~\cite{augs20randaugment,muller2021trivialaugment}, and Mixup~\cite{zhang2017mixup}. 
FixMath~\cite{ssl20fixmatch} inherited the idea of strong augmentations and further boosted the accuracy by a fixed threshold to select confident pseudo-labels. 
More recent research intended to introduce additional training and supervision, like using 
% contrastive learning~\cite{ssl22lassl}, 
contrastive learning~\cite{ssl22lassl,ssl22cac}, 
% distribution alignment~\cite{ssl20remix}, 
distribution alignment~\cite{ssl20remix,ssl22dcssl}, 
and Sinkhorn-Knopp clustering~\cite{ssl21SLA}, to further enhance the performance.

Motivated by the progress in semi-supervised classification, some studies aim to achieve dense segmentation performance with only a fraction of labels. Generally, recent jobs can be categorized into two main groups. 1) rectifying the pseudo-labels by training extra correcting networks~\cite{sss20scn,sss20ecs,sss22ELN}, re-balancing the classes~\cite{sss21dars}, or using multiple predictions~\cite{sss22PSMT}; 2) exploring more supervisions by using extra losses~\cite{sss21cps}, utilizing stronger augmentations~\cite{sss22st++,sss21simple}, or applying the advanced contrastive learning~\cite{sss22u2pl,sss21pc2seg,sss21semisegcontrast,sss21c3seg}. These studies show promising results at the cost of integrating extra network components or additional training processes. To the best of our knowledge, all the existing studies indiscriminately perturb unlabeled samples and minimize an average consistency loss over all unlabeled samples. Differently, we differentiate  different samples in terms of the learning difficulty, evaluated as instance hardness. We utilize the hardness to guide the training process and achieve new SOTA performance on several semi-supervised semantic segmentation benchmarks.

Instance hardness~\cite{smith2014instance,prudencio2015analysis,smith2016comparative,chang2017active} has been widely studied in hard example mining~\cite{yuan2017hard} and curriculum learning~\cite{zhou2020curriculum}. 
Their evaluation mainly depends on the instantaneous or historical training losses with respect to ground truths. 
Lacking accurate label information makes hardness measurements of unlabeled instances much more challenging. 
Some works~\cite{yuan2017hard,jin2018unsupervised,wang2020instance} perform hardness analysis on unlabeled data to split all the samples into the hard and easy groups by sorting or ranking the hardness with a predefined threshold. Such methods only require \textbf{qualitative} analysis for selecting or filtering purposes. However, specific \textbf{quantitative} hardness analysis, especially on segmentation tasks, is still under-explored. In iMAS, we need the quantitative hardness to determine the mixup between strongly and weakly augmented crops, as well as the exact unsupervised loss weight for each unlabeled instance. Thus we propose a new class-weighted symmetric metric to evaluate the hardness of unlabeled instances in segmentation tasks.

%%%%%%%%%%%%%%%%%%%%%%%%%%%%%%%%%%%%
%%  3. methods
%%%%%%%%%%%%%%%%%%%%%%%%%%%%%%%%%%%%
\section{Method}
\label{sec:method}
The goal of semi-supervised semantic segmentation is to generalize a segmentation model by effectively leveraging a labeled training set $D_x=\{(x_i, y_i)\}_{i=1}^{|D_x|}$ and a large unlabeled training set $D_u=\{u_i\}_{i=1}^{|D_u|}$, with typically $|D_x|\ll |D_u|$. In our method, following the consistency regularization (CR) based semi-supervised classification approaches~\cite{ssl20fixmatch,ssl20uda}, we aim to train the segmentation encoder and decoder on both labeled and unlabeled data simultaneously. In each iteration, given a batch of labeled samples $\mathcal{B}_x=\{(x_i, y_i)\}_{i=1}^{|\mathcal{B}_x|}$ and unlabeled samples $\mathcal{B}_u=\{u_i\}_{i=1}^{|\mathcal{B}_u|}$, the overall training loss is formulated as,
\begin{align}
    \label{equ:loss:total}
    \mathcal{L} = \mathcal{L}_x + \lambda_u \mathcal{L}_u,
\end{align}
where $\lambda_u$ is a scalar hyper-parameter to adjust the relative importance between the supervised loss $\mathcal{L}_x$ on $\mathcal{B}_x$ and the unsupervised loss $\mathcal{L}_u$ on $\mathcal{B}_u$. 
Without introducing extra losses or network components, iMAS can perform effectively quantitative hardness analysis for each unlabeled instance and then supervise the training on unlabeled data in a model-adaptive fashion across the training course.
% Without introducing extra losses or network components, iMAS can effectively evaluate the instance hardness and then supervise the training on unlabeled data in a model-adaptive fashion across the training course. 
In this section, we first introduce our proposed iMAS at a high level in Sec.~\ref{method:overview} and then present the detailed designs in terms of the quantitative hardness analysis in Sec.~\ref{method:eval} and the model-adaptive guidance in Sec.~\ref{method:hag}.

\begin{algorithm*}[t]
\caption{iMAS algorithm in a mini-batch.}
\label{alg:algorithm}
\textbf{Input}: Labeled batch $\mathcal{B}_x=\{(x_i, y_i)\}_{i=1}^{|\mathcal{B}_x|}$, unlabeled batch $\mathcal{B}_u=\{u_i\}_{i=1}^{|\mathcal{B}_u|}$ ($|\mathcal{B}_x|=|\mathcal{B}_u|$), hardness evaluation strategy $\phi$, weak augmentation $\mathcal{A}_w(\cdot)$, adaptive strong augmentation $\mathcal{A}_s(\cdot)$\\
\textbf{Parameter}: confidence threshold $\tau$,  unsupervised loss weight $\lambda_{u}$
\begin{algorithmic}[1] 
\STATE $\mathcal{L}_x = \frac{1}{|\mathcal{B}_x|} \sum_{i=1}^{|\mathcal{B}_x|} \frac{1}{H\times W}\sum_{j=1}^{H\times W} \mathrm{H}(\hat{y}_i(j), y_i(j))$ \quad \textcolor{gray}{//    calculate the supervised loss.}
\FOR{$u_i \in \mathcal{B}_u$} 
    \STATE $p^s_i = f_{\theta_s} (\mathcal{A}_w(u_i))$ \quad \textcolor{gray}{//    obtain segmentation predictions on weakly-augmented instances.}
    \STATE $p^t_i = f_{\theta_t} (\mathcal{A}_w(u_i))$ \quad \textcolor{gray}{//    obtain pseudo-labels from the teacher model.}
    \STATE $\gamma_i = \phi(p^t_i, p^s_i)$ \quad   \textcolor{gray}{//   evaluate the hardness of each instance.}
\ENDFOR
\STATE $\mathcal{L}_u = \frac{1}{|\mathcal{B}_u|} \sum_{i=1}^{|\mathcal{B}_u|}  \frac{\gamma_i}{2 H\times W}\sum_{j=1}^{H\times W} [\mathbbm{1}(\max(p^t_i(j)) \geq \tau) \mathrm{H}(f_{\theta_s}(\mathcal{A}_s^I(u_i)), p^t_i(j)) + $ \\ \, $\mathbbm{1}(\max(p^{t'}_i(j)) \geq \tau) \mathrm{H}(f_{\theta_s}(\mathcal{A}_s^C(u_i)), p^{t'}_i(j))]$ \quad \textcolor{gray}{//    calculate model-adaptive consistency loss}
\RETURN $\mathcal{L} = \mathcal{L}_x + \lambda_u \mathcal{L}_u$
\end{algorithmic}
\end{algorithm*}

\subsection{Overview}
\label{method:overview}

Built on top of the CR-based semi-supervised framework, iMAS jointly trains a student model with learnable weights $\theta_s$ and a teacher model with learnable weights $\theta_t$ in a mutually-beneficial manner. The complete algorithm is shown in  algorithm~\ref{alg:algorithm}. On the one hand, the teacher model is updated by the exponential moving averaging of the student weights, \textit{i.e.,} 
\begin{align}
    \theta_t \leftarrow \alpha \theta_t + (1 - \alpha) \theta_s,
\end{align}
where $\alpha$ is a common momentum parameter, set as 0.996 by default. On the other hand, the student model relies on the pseudo-labels generated by the teacher model to be trained on the unlabeled data. Specifically, the student model is trained via minimizing the total loss $\mathcal{L}$ in Equation~\ref{equ:loss:total}, which consists of two cross-entropy loss terms, $\mathcal{L}_u$ and $\mathcal{L}_x$, applied on labeled and unlabeled data, respectively. Let $\mathrm{H}(z_1, z_2)$ denote the cross-entropy loss between prediction distributions $z_1$ and $z_2$. The supervised loss $\mathcal{L}_x$ is calculated as,
\begin{align}
    \mathcal{L}_x = \frac{1}{|\mathcal{B}_x|} \sum_{i=1}^{|\mathcal{B}_x|} \frac{1}{H\times W}\sum_{j=1}^{H\times W} \mathrm{H}(\hat{y}_i(j), y_i(j)),
\end{align}
where $\hat{y}_i = f_{\theta_s} (\mathcal{A}_w(x_i))$, represents the segmentation result of the student model on the $i$-th weakly-augmented labeled instance. 
$j$ represents the $j$-th pixel on the image or the corresponding segmentation mask with a resolution of $H\times W$. The weak augmentation $\mathcal{A}_w$ includes standard resizing, cropping, and flipping operations. Importantly, the way to leverage the unlabeled data is the key to semi-supervised learning and also the crucial part differentiating our method from others. In most CR-based studies, the standard (${std}$) unsupervised loss $\mathcal{L}_u^{std}$ is simply,
\begin{align}
    \mathcal{L}_{u}^{std}\!=\!\frac{1}{|\mathcal{B}_u|}\!\sum_{i=1}^{|\mathcal{B}_u|}\!\frac{1}{H\times W}\sum_{j=1}^{H\times W}\!\mathbbm{1}(\max(p^t_i(j))\!\geq\!\tau)\!\mathrm{H}(\hat{p}_i(j),\!p^t_i(j)),
\end{align}
where $\hat{p}_i=f_{\theta_s}(\mathcal{A}_s^{std}(u_i))$ represents
the segmentation output of the student model on the $i$-th unlabeled instance augmented by $\mathcal{A}_s^{std}$, while $p^t_i=f_{\theta_t}(\mathcal{A}_w(u_i))$ represents the segmentation outputs of the teacher model on the $i$-th weakly-augmented unlabeled instance.
$\tau$ is a predefined confidence threshold to select high-confidence predictions.
$\mathcal{A}_s^{std}$ represents standard \textbf{instance-agnostic} strong augmentations, including intensity-based data augmentations~\cite{augs20randaugment} and CutMix~\cite{augs19cutmix} as shown in Table~\ref{tab:augs}. However, such operations are limited in ignoring the differences and learning difficulties among unlabeled samples. 

Differently, in our iMAS, we treat each instance discriminatively and provide instance-specific supervision on the training of unlabeled data. As shown in Figure~\ref{fig:diagram}, we first evaluate the hardness of each weakly-augmented unlabeled instance via strategy $\phi$, and then employ the \textbf{instance-specific and model-adaptive} supervision on the strong augmentations $\mathcal{A}_s$ as well as the calculations of unsupervised loss $\mathcal{L}_u$, which are elaborated in following sections.

% \subsection{Hardness evaluation of unlabeled instances}
\subsection{Quantitative hardness analysis}
\label{method:eval}
In iMAS, we perform quantitative hardness analysis to differentiate distinct unlabeled samples. In most hardness-related studies, the instantaneous or historical training losses~\cite{zhou2020curriculum,smith2014instance} to the ground truth are used to assess the instance hardness. 
However, in semi-supervised segmentation, evaluating the hardness of unlabeled data is challenging at 1) lacking accurate ground-truth labels and 2) dynamic changes closely related to the model performance. A ``hard" sample can become easier with the evolution of the model, but such dynamics cannot be easily identified without accurate label information. Inspired by \cite{grill2020bootstrap,wang2020instance}, it is more difficult for the teacher and student models to achieve consensus on a hard instance. Hence we design a symmetric class-weighted IoU between the segmentation results of the student and teacher models to evaluate the instantaneous hardness. The class-weighted design is used to alleviate the class-imbalanced issue in segmentation tasks. 

Such evaluation, denoted by $\phi$, can be regarded as a function of the model performance and dynamically estimate the training difficulties of unlabeled crops throughout the training process. Specifically, as shown in Figure~\ref{fig:diagram}, we first obtain the segmentation predictions $p^s_i$ and $p^t_i$ on the $i$-th weakly-augmented unlabeled instance, from the student and teacher models, respectively,
\begin{align}
    p^s_i&\!=\!f_{\theta_s}\!(\mathcal{A}_w(u_i)\!), \rho_i^s\!=\!\frac{1}{H\!\times\!W}\!\sum_{j\!=\!1}^{H\!\times\!W}\!\mathbbm{1}\!(\max(p^s_i(j)\!)\!\ge\!\tau\!) \\
    p^t_i&\!=\!f_{\theta_t}\!(\mathcal{A}_w(u_i)\!), \rho_i^t\!=\!\frac{1}{H\!\times\!W}\!\sum_{j\!=\!1}^{H\!\times\!W}\!\mathbbm{1}\!(\max(p^t_i(j)\!)\!\ge\!\tau\!)
\end{align}
where $\rho_i^s$ and $\rho_i^t$ represent the high-confidence ratios on $p^s_i$ and $p^t_i$, respectively.
Let $\mbox{wIOU}(z_1, z_2)$ denote the class-weighted IoU between segmentation predictions $z_1$ and $z_2$. Note that, this evaluation is not commutative, \textit{i.e.,} $\mbox{wIOU}(z_1, z_2) \neq \mbox{wIOU}(z_2, z_1)$. To make wIoU valid for hardness evaluation at each iteration, the symmetric hardness $\gamma_i$ for $i$-th unlabeled instance is calculated as,
\begin{align}
    \gamma_i\!=\!\phi(p^t_i,\!p^s_i)\!=\!1\!-\![\frac{\rho_i^s}{2} \mbox{wIOU}(p^s_i, p^t_i) + \frac{\rho_i^t}{2} \mbox{wIOU}(p^t_i, p^s_i)]
\end{align}
where $1/2$ ensures the hardness is in $[0,1]$. In this way, the harder instance that requires better generalization ability holds a larger value of $\gamma$ while the easier one will be identified by a smaller $\gamma$. % jimin：hard -> harder 

\subsection{Model-adaptive supervision}
\label{method:hag}
With the quantitative hardness evaluation for each unlabeled instance, we carefully inject such information into the training process by instance-specific and model-adaptive strong perturbations and loss modifications. 
Specifically, we first leverage the instance hardness for adaptive augmentations both individually and mutually.
By ``individually", we adjust the intensity-based augmentation applied on each instance according to its absolute hardness value; by ``mutually", we replace random pairs of unlabeled data in CutMix with specific \textbf{hard-easy pairs} assigned by sorting the corresponding hardness. 
Moreover, instead of indiscriminately averaging the losses, we \textbf{weigh} the losses of different unlabeled instances by multiplying their corresponding hardness. We present these details below.

\subsubsection{Model-adaptive strong augmentations}
The popular strong augmentations in recent semi-supervised segmentation studies mainly consist of two different types: intensity-based augmentation and CutMix, as shown in Table~\ref{tab:augs}. In iMAS, we apply instance-specific adjustments to both types of augmentations.

\textbf{Intensity-based augmentations}.
Standard intensity-based data augmentations randomly select two kinds of image operations from an augmentation pool and apply them to the weakly-augmented instances. 
However, as discussed by \cite{sss21simple}, strong augmentations may hurt the data distribution and degrade the segmentation performance, especially during the early training phase. 
Unlike distribution-specific designs~\cite{sss21simple}, we simply adjust the augmentation degree for an unlabeled instance by mixing its strongly-augmented and weakly-augmented outputs. 
Formally, the ultimate augmented output of the $i$-th unlabeled instance, $\mathcal{A}_s^{I}(u_i)$, can be obtained by,
\begin{align}
    \mathcal{A}_s^{I} (u_i) \leftarrow \gamma_i \mathcal{A}_s^{I} (u_i) + (1 - \gamma_i) \mathcal{A}_w(u_i)
\end{align}
where the distortion caused by the intensity-based strong augmentation is proportionally weakened by the corresponding weakly-augmented output. In this way, harder instances with larger hardness are not perturbed significantly so that the model will not be challenged on potentially out-of-distribution cases. On the other hand, easier instances with lower values of $\gamma$, which have been well fitted by the model, can be further learned from their strongly-augmented variants. Such model-adaptive augmentations can better adjust to the model's generalization ability.

\textbf{CutMix-based augmentations}.
CutMix~\cite{augs19cutmix} is a widely adopted technique to boost semi-supervised semantic segmentation. 
It is applied between unlabeled instances with a predefined probability. 
It can randomly copy a region from one instance to another, and so do their corresponding segmentation results. 
The augmentation pairs are generated randomly.
Differently, in iMAS, we improve the standard CutMix by a model-adaptive design, which is distinct in two ways:
\textbf{1)} the mean hardness determines the trigger probability of CutMix augmentation over the mini-batch instead of using a predefined hyper-parameter;
\textbf{2)} the copy-and-paste pairs are assigned specifically between the hard and easy samples. 
According to the instance hardness, we obtain two sequences by sorting unlabeled samples of a mini-batch in the ascending and descending orders, respectively. 
We then aggregate two sequences element-by-element to generate the hard-easy pairs.
Formally, given a specific hard-easy pair, $(u_m, u_n)$, the model-adaptive CutMix can be expressed as,
\begin{align}
\left.\begin{aligned}
    \mathcal{A}_s^{C} (u_m) &\leftarrow M_m \odot u_n + (\mathbf{1} - M_m) \odot u_m   \\
    p^{t'}_m &\leftarrow M_m \odot p^t_n + (\mathbf{1} - M_m) \odot p^t, \\
    \mathcal{A}_s^{C} (u_n) &\leftarrow M_n \odot u_m + (\mathbf{1} - M_n) \odot u_n \\
    p^{t'}_n &\leftarrow M_n \odot p^t_m + (\mathbf{1} - M_n) \odot p^t_n \\
\end{aligned}\right\}, \\
\mbox{by a triggering probability of } \overline{\gamma} = \frac{1}{|\mathcal{B}_u|}\sum_{n=1}^{|\mathcal{B}_u|} \gamma_n,
\end{align}
where $M_m$ and $M_n$ denote the randomly generated region masks for $u_m$ and $u_n$, respectively.  Besides, the pseudo-labels need to be revised accordingly after applying CutMix data augmentations, obtaining $p^{t'}_m$ and $p^{t'}_n$. 
This mutual augmentation is applied following a Bernoulli process, \textit{i.e.,} triggered only when a uniformly random probability is higher than the average hardness $\overline{\gamma}$.

\begin{table}[t]
\centering
\resizebox{0.99\linewidth}{!}{
\begin{small}
\begin{tabular}{ll}
\toprule
\multicolumn{2}{c}{\textbf{Weak} Augmentations} \\
\midrule
Random scale & Randomly resizes the image by $[0.5, 2.0]$. \\
Random flip & Horizontally flip the image with a probability of 0.5. \\
Random crop & Randomly crops an region from the image ($513\times513$, $769\times769$). \\
\midrule
% \midrule
\multicolumn{2}{c}{\textbf{Strong} intensity-based Augmentations} \\
\midrule
Identity & Returns the original image. \\
Invert & Inverts the pixels of the image.\\
Autocontrast & Maximizes (normalize) the image contrast.\\
Equalize & Equalize the image histogram.\\
Gaussian blur & Blurs the image with a Gaussian kernel.\\
Contrast & Adjusts the contrast of the image by [0.05, 0.95]. \\
Sharpness & Adjusts the sharpness of the image by [0.05, 0.95]. \\
Color & Enhances the color balance of the image by [0.05, 0.95] \\
Brightness & Adjusts the brightness of the image by [0.05, 0.95] \\
Hue & Jitters the hue of the image by [0.0, 0.5] \\
Posterize & Reduces each pixel to [4,8] bits.\\
Solarize & Inverts all pixels of the image above a threshold value from [1,256).\\
\midrule
% \midrule
\multicolumn{2}{c}{\textbf{CutMix} augmentation} \\
\midrule
CutMix & Copy and paste random size regions among different unlabeled images. \\
\bottomrule
\end{tabular}
\end{small}
}
\caption{List of various image transformations in iMAS.}
\label{tab:augs}
\vspace{-1em}
\end{table}

\begin{table*}[t]
  \centering
  % \resizebox{0.99\linewidth}{!}{
  \begin{tabular}{r|ccc|ccc}
    \toprule 
    \multirow{2}{*}{Method} &\multicolumn{3}{c|}{ResNet-$50$} & \multicolumn{3}{c}{ResNet-$101$} \\
    \cline{2-7}
    % \cmidrule(r){2-5} \cmidrule(r){6-9} 
    & $1/16$ ($662$)   & $1/8$ ($1323$)  & $1/4$ ($2646$)  & $1/16$ ($662$)  & $1/8$ ($1323$)  & $1/4$ ($2646$) \\
    % \hline
    \midrule
    Supervised$^*$ & $63.8$ & $69.0$ & $72.5$ & $67.4$ & $72.1$ & $74.7$\\
    % \hline
    % \midrule
    MT~\cite{ssl17mt}   & $66.8$ & $70.8$ & $73.2$ & $70.6$ & $73.2$ & $76.6$ \\
    CCT~\cite{sss20cct} &  $65.2$ & $70.9$ & $73.4$ & $68.0$ & $73.0$ & $76.2$ \\
    CutMix-Seg~\cite{sss20cutmixseg} &  $68.9$ & $70.7$ & $72.5$ & $72.6$ & $72.7$ & $74.3$ \\
    GCT~\cite{sss20gct} & $64.1$ & $70.5$ & $73.5$ & $69.8$ & $73.3$ & $75.3$ \\
    CAC~\cite{sss21cac} & $70.1$ & $ 72.4$ & $74.0$ & $72.4$ & $74.6$ & $76.3$  \\
    CPS~\cite{sss21cps} & $72.0$ & $ 73.7$ & $74.9$ & $74.5$ & $76.4$ & $77.7$ \\ 
    PSMT\dag~\cite{sss22PSMT} &72.8 & 75.7 & 76.4 & 75.5 & 78.2 & 78.7 \\
    ELN~\cite{sss22ELN} & $70.5$ & $73.2$ & $74.6$  & $72.5$ & $75.1$ & $76.6$  \\
    ST++~\cite{sss22st++} & $72.6$ & $74.4$ & $75.4$  & $74.5$ & $76.3$ & $76.6$ \\
    \textbf{iMAS (ours)} & ${74.8}$ & ${76.5} $ & ${77.0}$ & ${76.5} $ & ${77.9}$ & ${78.1}$ \\
    % \hline
    \midrule
    U$^2$PL\ddag~\cite{sss22u2pl}  & $72.0$ & $ 75.2 $ & $76.2 $ & $74.4$ & $77.6$ & $78.7$ \\ 
    %https://github.com/Haochen-Wang409/U2PL/issues/3
    \rowcolor{blue!30} \textbf{iMAS (ours)}\ddag & $\bf{75.9}$ & $\bf{76.7} $ & $\bf{77.1} $ & $\bf{77.2}$ & $\bf{78.4}$ & $\bf{79.3}$ \\ 
    \bottomrule
\end{tabular}
% }
\caption{Comparison with SOTA methods on  \textbf{PASCAL VOC 2012} \texttt{val} set under different partition protocols. Labeled images are sampled from the \textit{blender} training set (augmented by SBD dataset), including $10,583$ samples in total.
  \ddag\ means the results are obtained by setting the output\_stride as 8 in DeepLabV3$+$ (16 for others). 
  % \dag\ means running more epochs in PSMT.
  $^*$ denotes our reproduced results. Best results  are highlighted in \textbf{bold}. }
  \label{tab:voc:blend}
\end{table*}

\subsubsection{Model-adaptive unsupervised loss}
Considering the learning difficulty of each instance, we design a model-adaptive unsupervised loss to learn from unlabeled data differentially. 
Inspired by curriculum learning~\cite{bengio2009curriculum}, we prioritize the training on easy samples over hard ones. 
Precisely, we weigh the unsupervised losses for each instance by multiplying their corresponding easiness, evaluated by $1 - \gamma$. 
% Since the easier samples are assigned with higher hardness values in our method, we directly weigh the unsupervised losses for each instance by multiplying their corresponding hardness evaluations. 
Combined with model-adaptive augmentations, we can calculate the unsupervised loss by,
\begin{equation}
    \begin{split}
        &\mathcal{L}_u\!=\!\frac{1}{|\mathcal{B}_u|} \sum_{i=1}^{|\mathcal{B}_u|}  \frac{1 - \gamma_i}{2 H\times W}\sum_{j=1}^{H\times W} [\mathbbm{1}(\max(p^t_i(j)) \geq \tau) \mathrm{H}(f_{\theta_s}\\&(\mathcal{A}_s^I(u_i)), p^t_i(j))\!+\!\mathbbm{1}(\max(p^{t'}_i(j))\!\geq\!\tau) \mathrm{H}(f_{\theta_s}(\mathcal{A}_s^C(u_i)), p^{t'}_i(j))].
    \end{split}
\end{equation}
Since the hardness is evaluated upon each (weakly augmented) image instance, under its guidance, the two strong augmentations are performed separately rather than in a cascading manner.
In this way, the model will not be trained on over-distorted variants, and our model-adaptive designs can be effectively utilized.

%%%%%%%%%%%%%%%%%%%%%%%%%%%%%%%%%%%%
%%  4. Experiments
%%%%%%%%%%%%%%%%%%%%%%%%%%%%%%%%%%%%

\section{Experiments}
\label{sec:exps}

In this section, we examine the efficacy of our method on standard semi-supervised semantic segmentation benchmarks and conduct extensive ablation studies to further verify the superiority and stability.

\textbf{Dataset and backbone}. Following recent SOTAs~\cite{sss21cps,sss22st++} in semi-supervised segmentation, we adopt DeepLabv3$+$~\cite{seg18deeplabv3plus} based on Resnet~\cite{he16resnet} as our segmentation backbone and investigate the test performance on Pascal VOC2012~\cite{data15voc} and Cityscapes~\cite{data16citys}, in terms of the mean intersection-over-union (mIOU). The classical VOC2012 consists of 21 classes with 1464 training and 1449 validation images. As a common practice,  the blended training set is also involved, including additional 9118 training images from the Segmentation Boundary (SBD) dataset~\cite{data11vocaugs}. Cityscapes is a large dataset on urban street scenes with 19 segmentation classes. It consists of 2975 training and 500 validation images with fine annotations.

\textbf{Implementation details}. For both the student and the teacher models, we load the ResNet weights pre-trained on ImageNet~\cite{data09imagenet} for the encoder and randomly initialize the decoder. An SGD optimizer with a momentum of 0.9 and a polynomial learning-rate decay with an initial value of 0.01 are adopted to train the student model. The total training epoch is 80 for VOC2012 and 240 for Cityscapes. Following~\cite{sss22u2pl}, training images are randomly cropped into $513\times513$ and $769\times769$ for Pascal VOC2012 and Cityscapes, respectively.
On Cityscapes, we also use the sliding evaluation to examine the performance on validation images with a resolution of $1024\times2048$. We set $\mathcal{B}_u = \mathcal{B}_x = 16$ and adopt the sync-BN for all runs.

\begin{table}[t]
\centering
\resizebox{0.99\linewidth}{!}{
\begin{tabular}{r|ccccc}
\toprule
\multirow{2}{*}{Method} & 
1/16 & 1/8 & 1/4  & 1/2 & Full  \\
& (92) & (183) & (366) & (732) & (1464) \\
\midrule
Supervised $^*$ & 
45.5 & 57.5 & 66.6 & 70.4 & 72.9 \\
% \midrule
% MT~\cite{ssl17mt} &  51.7 & 58.9 & 63.9 & 69.5 & 71.0 \\
CutMix-Seg~\cite{sss20cutmixseg} & 
52.2 & 63.5 & 69.5 & 73.7 & 76.5 \\
PseudoSeg~\cite{sss20pseudoseg} & 
57.6 & 65.5 & 69.1 & 72.4 & 73.2 \\
PC${}^2$Seg~\cite{sss21pc2seg} & 
57.0 & 66.3 & 69.8 & 73.1 & 74.2 \\
CPS~\cite{sss21cps} & 
64.1 & 67.4 & 71.7 & 75.9 & - \\
PSMT~\cite{sss22PSMT} & 65.8 & 69.6 & 76.6 & 78.4& 80.0 \\
ST++~\cite{sss22st++} & 
65.2 & 71.0 & 74.6 & 77.3 & 79.1 \\
\textbf{iMAS (ours)} & $68.8$ & $74.4$ & $78.5$ & $79.5$ & $81.2$\\
\midrule
U$^2$PL$\ddag$~\cite{sss22u2pl}  & 
68.0 & 69.2 & 73.7 & 76.2 & 79.5 \\
\rowcolor{blue!30}\textbf{iMAS\ddag (ours)}  & $\bf{70.0}$ & $\bf{75.3}$ & $\bf{79.1}$ & $\bf{80.2}$ & $\bf{82.0}$\\
% \footnotesize{(\textcolor{blue}{$+$15.82})
\bottomrule
\end{tabular}
}
\caption{
Comparison with SOTA methods on \textit{classic} \textbf{PASCAL VOC 2012} \texttt{val} set under different partition protocols. Labeled images are sampled from the official VOC \texttt{train} set, including $1,464$ samples in total. Results are reported using Resnet-101. All notations are the same as in Table \ref{tab:voc:blend}.
}
\label{tab:voc:classic}
\end{table}

\begin{table}[t]
\centering
% \resizebox{0.99\linewidth}{!}{
\begin{tabular}{r|cccc}
\toprule
\multirow{2}{*}{Method} & 
1/16 & 1/8  & 1/4  & 1/2 \\
& (186) & (372) &  (744) & (1488) \\
\midrule
Supervised $^*$ & 
64.0 & 69.2 & 73.0 & 76.4 \\
% \midrule
MT~\cite{ssl17mt} & 66.1 & 72.0 & 74.5 & 77.4 \\
%
% CutMix-Seg~\cite{sss20cutmixseg} & 67.1 & 65.82 & 68.33 & - \\
%
CCT~\cite{sss20cct} &  66.4 & 72.5 & 75.7 & 76.8 \\
GCT~\cite{sss20gct} & 65.8 & 71.3 & 75.3 & 77.1 \\
%
% CAC~\cite{sss21cac} & - & 69.7 & 72.7 & - \\
%
% CPS~\cite{sss21cps} & 69.8 & 74.4 & 76.9 & 78.6 \\
%
CPS~\cite{sss21cps} & 74.4 & 76.6 & 77.8 & 78.8 \\
CPS\dag ~\cite{sss22u2pl} & 69.8 & 74.3 & 74.6 & 76.8 \\
%
% PSMT~\cite{sss22PSMT} & - & 77.1 & 78.4 & 79.2 \\
PSMT~\cite{sss22PSMT} & - & 75.8 & 76.9 & 77.6 \\
ELN~\cite{sss22ELN}& - & 70.3 & 73.5 & 75.3 \\
ST++~\cite{sss22st++} &
- & 72.7 & 73.8 & - \\
% USRN~\cite{sss22USR} & 71.2 & 75.0 & - & - \\
U$^2$PL $^*$ \cite{sss22u2pl} &
67.8 & 72.5 & 74.8 & 77.1 \\
\textbf{iMAS (ours)} &  $74.3$ & $77.4$ & $78.1$ & $79.3$ \\
\midrule
U$^2$PL\ddag $^*$~\cite{sss22u2pl}  &  69.0 & 73.0 & 76.3 & 78.6 \\
\rowcolor{blue!30} \textbf{iMAS (ours)}\ddag &  $\bf{75.2}$ & $\bf{78.0}$ & $\bf{78.2}$ & $\bf{80.2}$ \\
% \footnotesize{(\textcolor{blue}{$+$3.24})}
\bottomrule
\end{tabular}
% }
\caption{Comparison with SOTA methods on \textbf{Cityscapes} \texttt{val} set under different partition protocols. Labeled images are sampled from the Cityscapes \texttt{train} set, including $2,975$ samples in total. Results are reported using Resnet-50. $^*$ and \dag\ represent reproduced
results in iMAS and U$^2$PL, respectively.
Results with \ddag\ are obtained by setting the output\_stride as 8 in DeepLabV3$+$. }
\label{tab:city}
\end{table}

\subsection{Comparison with State-of-the-Art Methods}
In this section, we demonstrate the superior performance of our iMAS on both classic and blended VOC 2012 and Cityscapes under different semi-supervised partition protocols.
It is noteworthy that, on blended VOC, U$^2$PL~\cite{sss22u2pl} %unfairly
prioritizes selecting high-quality labels from classic VOCs. Instead, we randomly sample labels from the entire dataset and adopt the same partitions as specified in~\cite{sss21cps,sss22PSMT}. Therefore, we reproduce corresponding results on U$^2$PL and evaluate iMAS with different output\_strides, 8 and 16, respectively, for fair comparisons.

\textbf{PASCAL VOC 2012}. In Tables~\ref{tab:voc:blend} and~\ref{tab:voc:classic}, we compare our iMAS with recent SOTA methods on blended and classic VOC, respectively. We can clearly see from Table~\ref{tab:voc:blend} that iMAS can consistently outperform others regardless of using ResNet-50 or ResNet-101 as the segmentation encoder. The performance gain becomes more noticeable and clear as fewer labels are available. \textit{e.g.}, in the 1/16 partition, iMAS can improve the supervised baseline by 11\% and 9.1\% when using ResNet-50 and ResNet-101 as the encoders, respectively, and improve the ST++~\cite{sss22st++} by 2.2\% and 2.0\%, accordingly. Checking the results among different partitions, we can also observe that iMAS can even obtain better performance while using fewer labels compared to other SOTAs. For example, iMAS can obtain a high mIOU of 75.9\% using only 662 labels, while U$^2$PL requires 1323 labels to obtain a comparable performance of 75.2\% mIOU on blended VOC.
It suggests our method is more label efficient and potentially a good solution for label-scarce scenarios. In classic VOC with high-quality labels, our methods can outperform SOTA methods by a notable margin, as shown in Table~\ref{tab:voc:classic}. We attribute this improvement to the model-adaptive guidance that treats each unlabeled instance differently and effectively leverages them by instance-specific strategies in HegSeg. Generally, in both classic and blended cases, reserving a large feature map (\textit{i.e.,} set output\_stride=8) can slightly improve the test performance. 

\begin{table}
  \centering
  \resizebox{0.95\linewidth}{!}{
  \begin{tabular}{c|c|c|c}
    \toprule
    \multicolumn{3}{c|}{iMAS on} & \multirow{2}{*}{ mIOU (\%)} \\
    \cline{1-3}
    % \hline
    Loss $\mathcal{L}_u$ & Augs of $\mathcal{A}_s^I$ & Augs of $\mathcal{A}_s^C$ \\
    % \midrule
    \hline
     &  &  & 72.1 \footnotesize{(\textcolor{blue}{supervised})} \\
    \checkmark &  &  & 75.5  \footnotesize{(\textcolor{blue}{3.4$\uparrow$})}\\
     & \checkmark &   &  76.5 \footnotesize{(\textcolor{blue}{4.4$\uparrow$})}\\
     &  & \checkmark &  76.9 \footnotesize{(\textcolor{blue}{4.8$\uparrow$})}\\
    \hline
    \checkmark & \checkmark & \checkmark & {\bf 77.9 \footnotesize{(\textcolor{blue}{5.8$\uparrow$})}}\\
    \bottomrule
  \end{tabular}
  }
\caption{Ablation studies on the effectiveness of the instance-specific model-adaptive supervision on the unsupervised loss, intensity-based and CutMix augmentations, respectively. Results are reported on \textbf{PASCAL VOC 2012} under the 1/8 (1323) partition using Resnet-101 as the backbone. Improvements over the supervised baseline are marked in \textcolor{blue}{blue}.}
  \label{tab:ab:component}
\end{table}

\begin{figure}[t]
    \centering
    \includegraphics[width=0.85\linewidth]{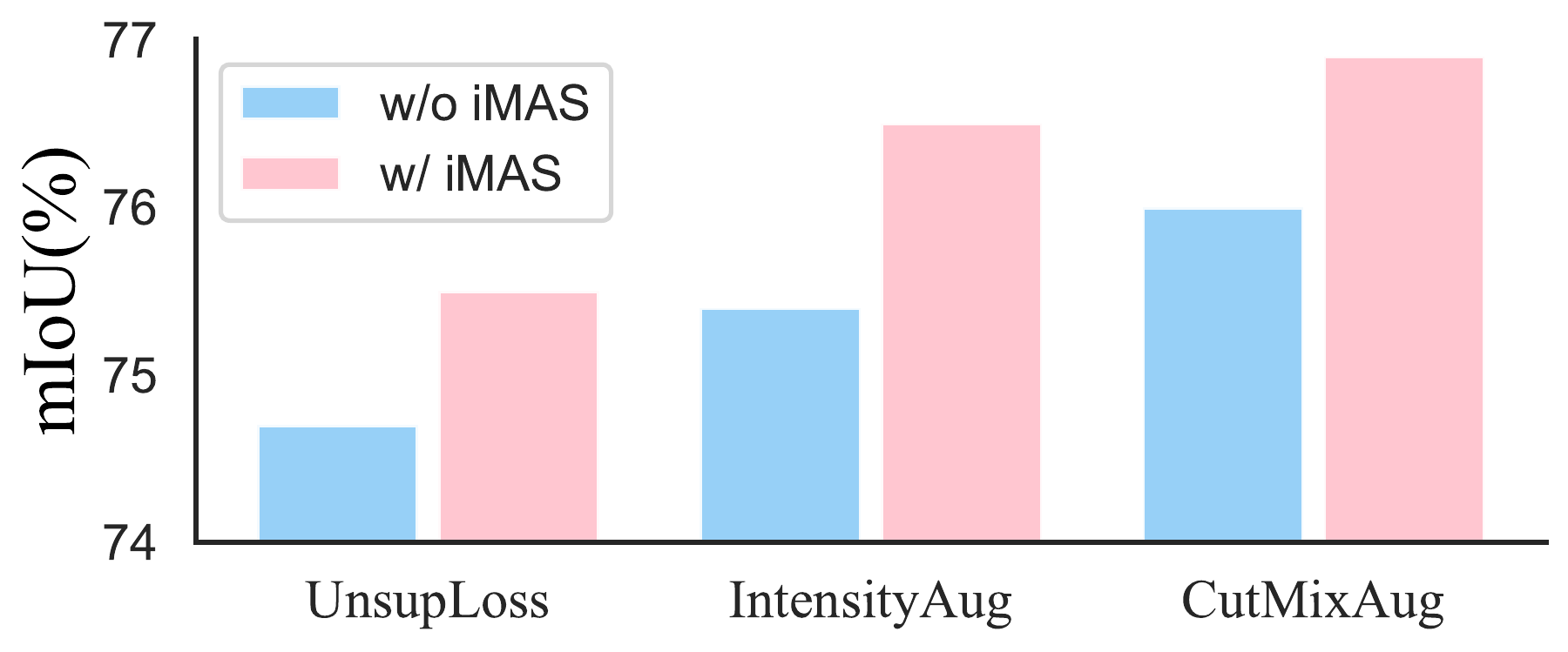}
    \caption{Effectiveness of iMAS on the unsupervised loss, intensity-based and CutMix augmentations, respectively.}
    \label{fig:abl:comp}
\end{figure}

\begin{figure*}
\centering
\begin{subfigure}{0.323\linewidth}
    \centering
    \includegraphics[width=0.99\linewidth]{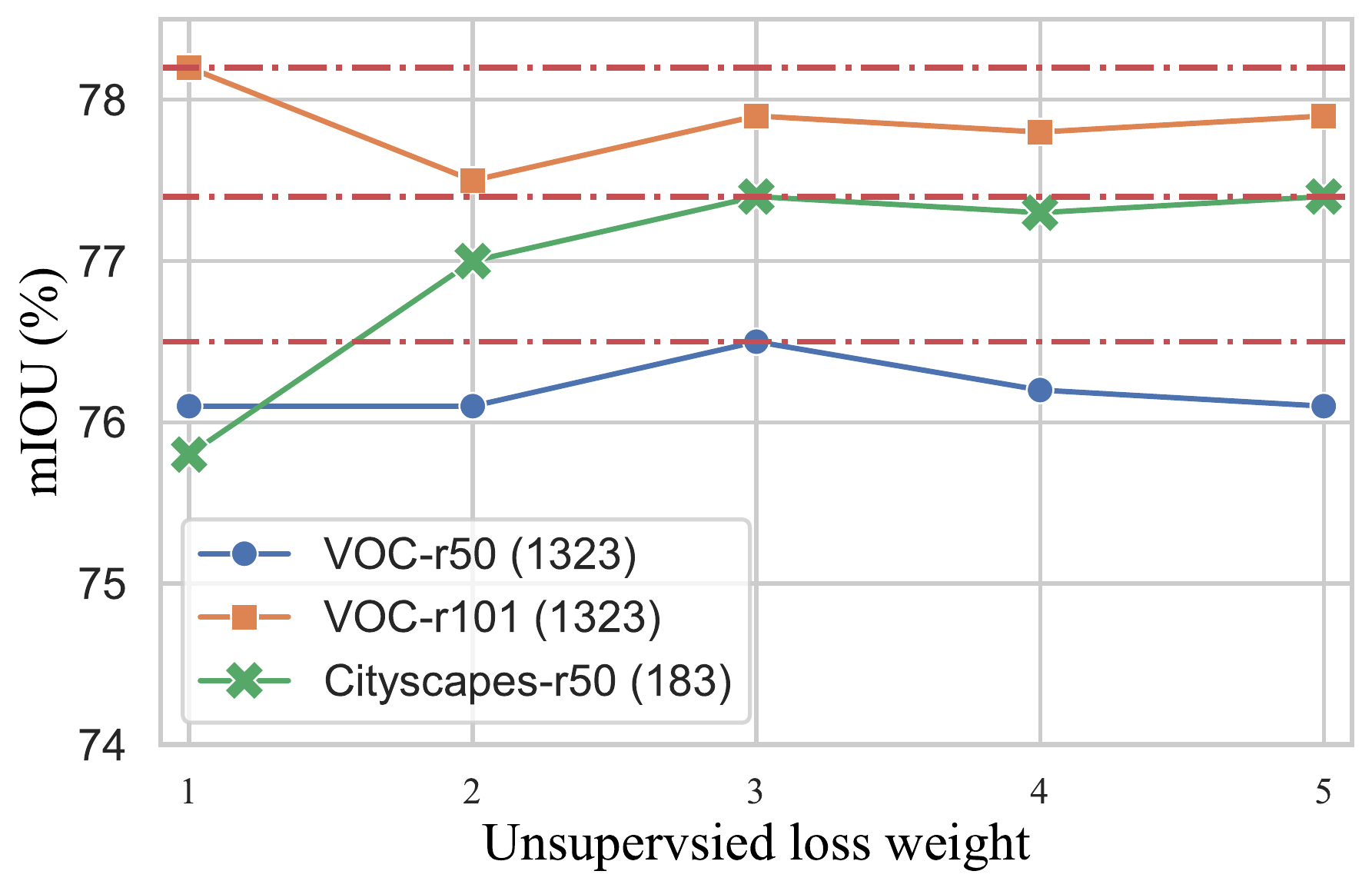}
    \caption{$\lambda_u$}
    \label{fig:abls:a}
\end{subfigure}
\hfill
\begin{subfigure}{0.323\linewidth}
    \centering
    \includegraphics[width=0.99\linewidth]{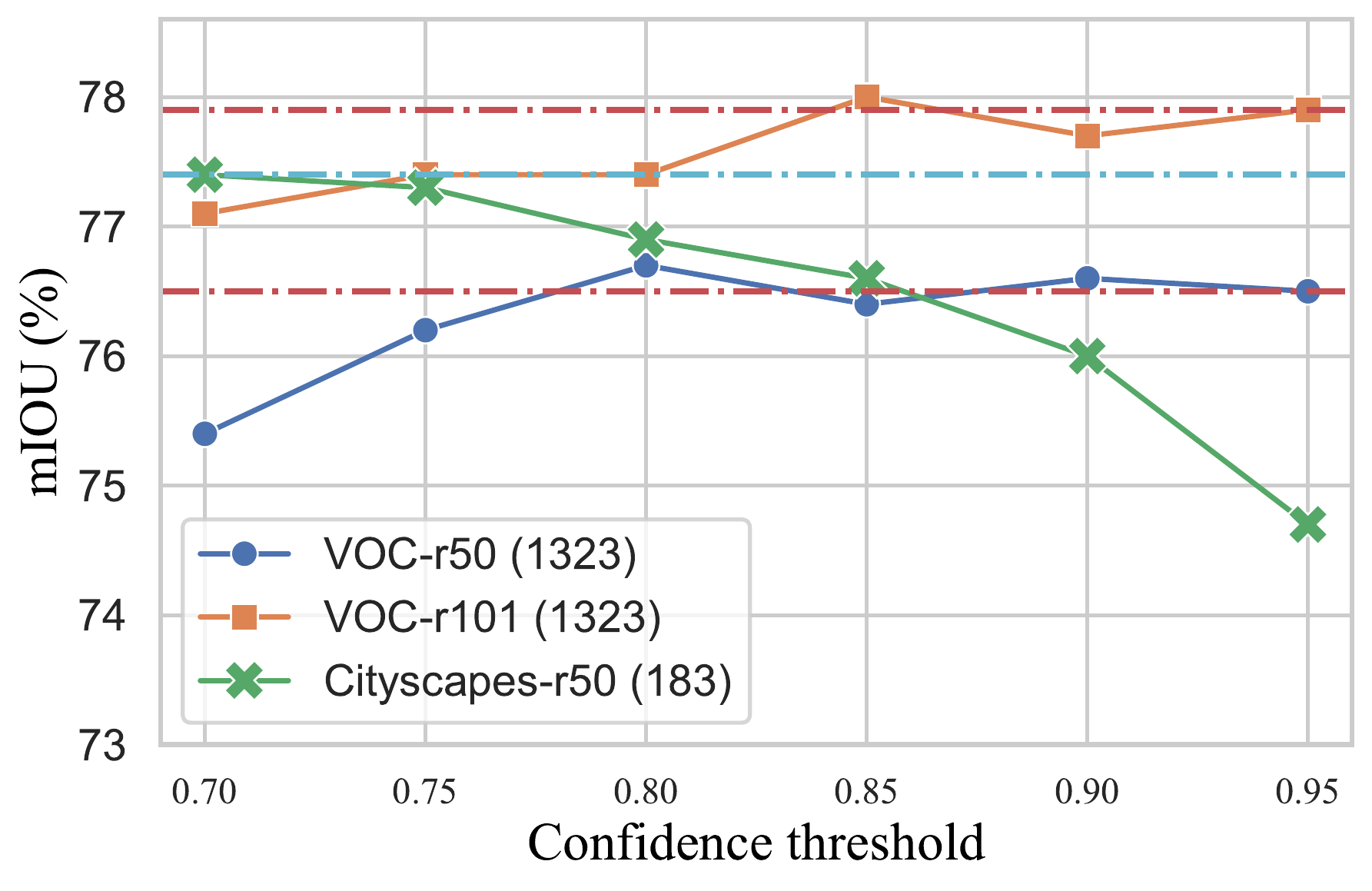}
    \caption{$\tau$}
    \label{fig:abls:b}
\end{subfigure}
\hfill
\begin{subfigure}{0.323\linewidth}
    \centering
    \includegraphics[width=0.99\linewidth]{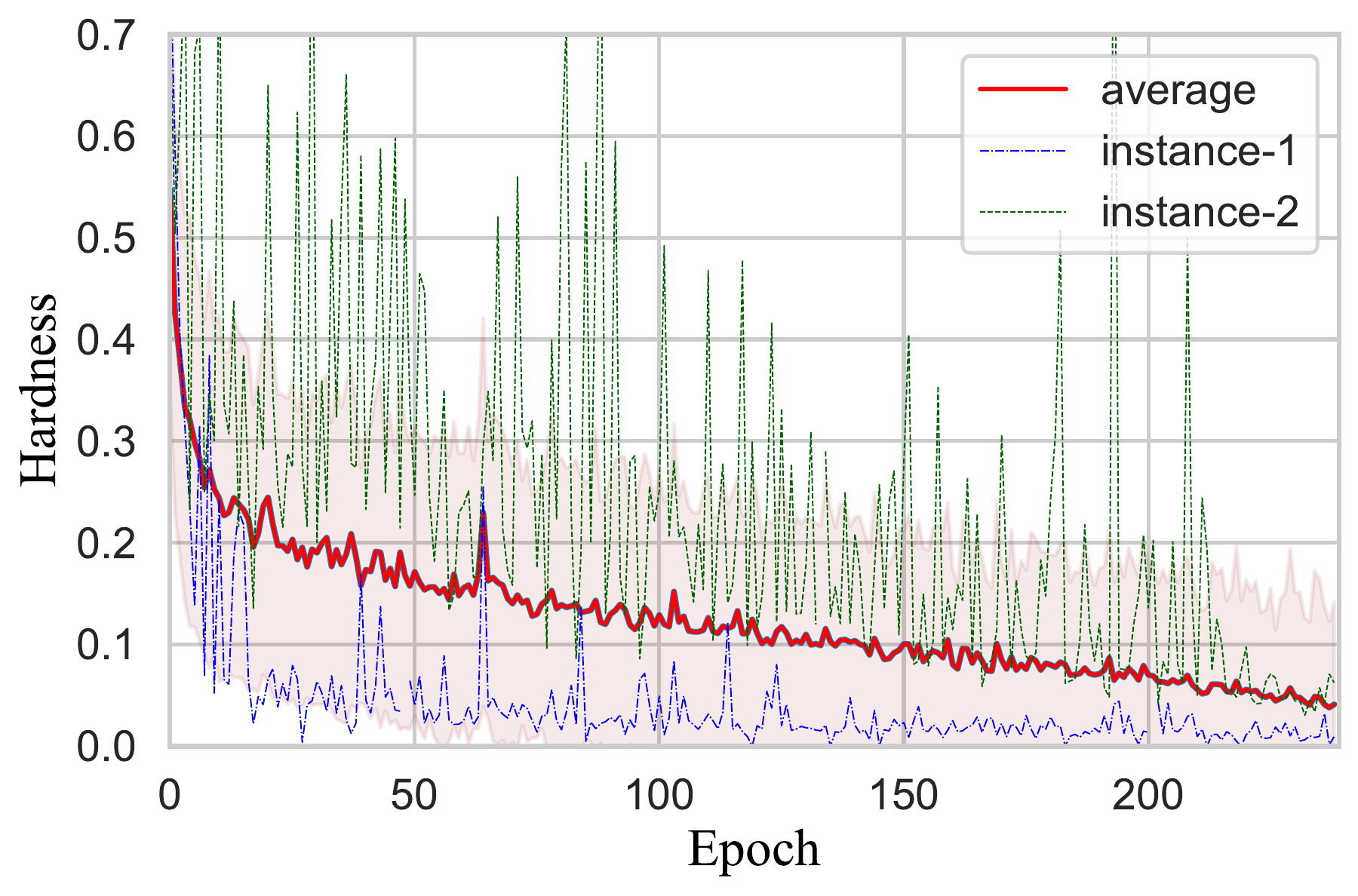}
    \caption{hardness}
    \label{fig:abls:c}
\end{subfigure}
\caption{We examine the effect of the loss weight and confidence threshold on VOC and Cityscapes under the 1/8 protocol in Figure (a) and (b), respectively. (c) shows how the mean instance hardness varies across the training course on Cityscapes under the 1/4 partition.} 
% Best viewed on screen. }
\label{fig:abls}
\end{figure*}

\textbf{Cityscapes}. In table~\ref{tab:city}, we evaluate our method on more challenging Cityscapes with ResNet-50 as the segmentation encoder. 
iMAS with output\_stride$=8$ can achieve high mIOUs of 75.2\%, 78.1\%, 78.2\%, 80.2\%,
in four different splits (1/16, 1/8, 1/4, 1/2), respectively. 
When output\_stride$=16$, given only 186 labeled images, iMAS can obtain a notable performance gain of 10.3\% against the supervised baseline and 6.5\%  against the previous best, U$^2$PL.
Not relying on any pseudo-rectifying networks~\cite{sss22ELN} or extra self-supervised supervisions~\cite{sss22u2pl}, iMAS achieves substantially better performance than the previous SOTAs, especially with fewer labels. 
Despite the simplicity of iMAS, the impressive performance further demonstrates the effectiveness and importance of our instance-specific and model-adaptive guidance. Surely, regardless of different semi-supervised approaches, we can see from Tables~\ref{tab:city} that providing more labeled samples can easily improve the semi-supervised performance.

\subsection{Ablations Studies}
We conduct ablation studies in the 1/8 partitions of blended VOC and Cityscapes, and examine the impact of the model-adaptive guidance and approach-related hyper-parameters.

\textbf{Effectiveness of model-adaptive guidance}. The key of iMAS lies in the instance-specific and model-adaptive guidance. In \cref{tab:ab:component}, we conduct a series of experiments on VOC2012 dataset to demonstrate its effectiveness on three components, the unsupervised loss, intensity-based and CutMix augmentations, respectively. 
It can been seen from \cref{fig:abl:comp} that performing model-adaptive guidance can consistently improve the standard operations, yielding around 1\% improvements on all standard counterparts. 
The powerfulness of strong augmentations can also be witnessed, as discussed in~\cite{sss22st++}. As a whole, iMAS can bring an improvement of 5.8\% against the supervised baseline.

\textbf{Impact of hyper-parameters}. In~\cref{fig:abls}, we investigate the influence of different $\lambda_u$ and $\tau$ on both datasets. It can be seen from \cref{fig:abls:a} that iMAS is not very sensitive to the loss weight on VOC while a large $\lambda_u$ is beneficial for Cityscapes. By default, we set $\lambda_u=3$ for all runs. According to \cref{fig:abls:b}, we set $\tau=0.95$ for VOC and $\tau=0.7$ for Cityscapes as default settings. This is simply because Cityscapes is a more challenging dataset requiring better discriminating ability and using a high-threshold will prevent models effectively learning from unlabeled samples. We can see from \cref{fig:abls:c} that both the mean and standard deviation of hardness evaluations on unlabeled data decrease as training processes and the model performance improves. Specifically, easy instances (\eg, Instance-1) can hold a low hardness from the very beginning, while the hardness of hard instances (\eg, Instance-2) fluctuates along the training process but eventually decreases.

\begin{figure}
    \centering
    \includegraphics[width=0.95\linewidth]{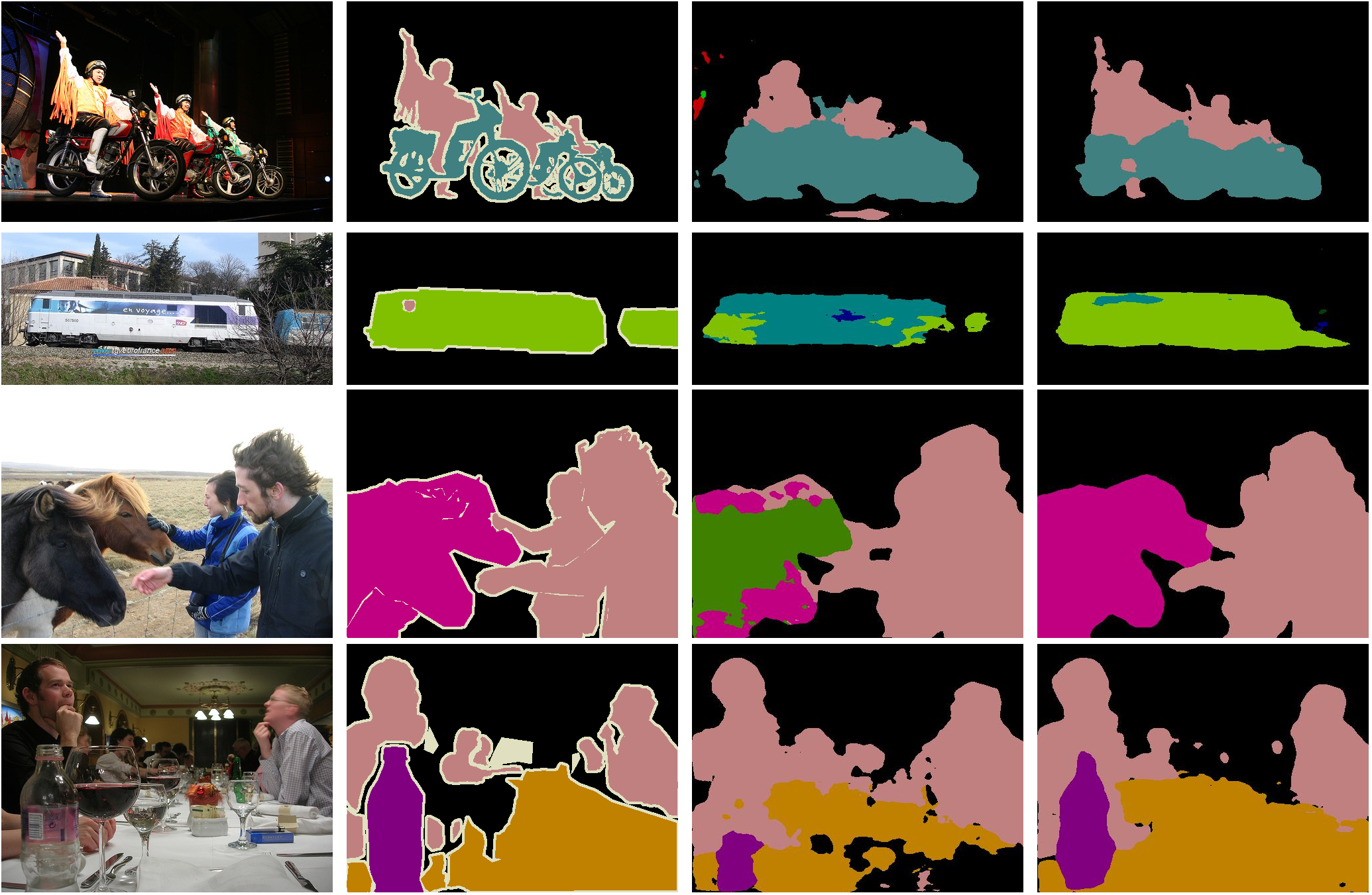}
    \caption{Qualitative results on Pascal VOC2012 using 183 fine labels. Columns from left to right denote the original images, the ground-truth, the supervised segmentation results, and the iMAS segmentation results, respectively.}
    \label{fig:abl:vis}
\end{figure}
\textbf{Qualitative Results}. We also present some segmentation results on Pascal VOC 2012 in Figure~\ref{fig:abl:vis} under the 183 partition protocol, using the Resnet-101 as the encoder. We can see that many mis-classified pixels and ignored segmentation details like arms in the supervised-only results are corrected in iMAS.

%%%%%%%%%%%%%%%%%%%%%%%%%%%%%%%%%%%%
%%  5. Conclusion
%%%%%%%%%%%%%%%%%%%%%%%%%%%%%%%%%%%%
\section{Conclusion}

In this paper, we highlight the instance uniqueness and propose iMAS, an instance-specific and model-adaptive supervision for semi-supervised semantic segmentation. Relying on our class-weighted symmetric hardness-evaluating strategies, iMAS treats each unlabeled instance discriminatively and employ model-adaptive augmentation and loss weighting strategies on each instance.
Without introducing additional networks or losses, iMAS can remarkably improve the SSS performance. 
Our method is currently \textbf{limited} at requiring forward unlabeled samples twice to obtain hardness evaluations. We hope iMAS can inspire SSS studies to explore more model-related and simple approaches.

\clearpage
%%%%%%%%% REFERENCES
{\small
\bibliographystyle{ieee_fullname}
\bibliography{imas}
}

\end{document}